\documentclass[letterpaper, 10 pt, conference]{ieeeconf}  

\IEEEoverridecommandlockouts                              

\usepackage{graphics} 
\usepackage{epsfig} 
\usepackage{times} 
\usepackage{amsmath} 
\usepackage{amssymb}  
\usepackage{xcolor}
\usepackage{mathtools}
\usepackage{multirow}
\makeatletter
\let\NAT@parse\undefined
\makeatother 
\usepackage[pagebackref=true,hidelinks]{hyperref}

\newcommand{\floor}[1]{\left\lfloor #1 \right\rfloor}

\newcommand{\seqset}{\mathcal{S}}
\newcommand{\set}[1]{\left\{{#1}\right\}}
\newcommand{\enorm}[1]{\left\|{#1}\right\|_2}
\DeclareMathOperator{\loss}{loss}
\DeclareMathOperator{\cploss}{CPloss}
\newcommand{\reals}[1]{\mathbb{R}^{#1}}
\newcommand{\binary}[1]{\mathbb{B}^{#1}}
\newcommand{\one}{\mathbf{1}}
\newcommand{\comment}[1]{}

\title{\LARGE \bf
Human Action Forecasting by Learning Task Grammars
}

\author{
Tengda Han
\and
Jue Wang
\and
Anoop Cherian
\and
Stephen Gould
\thanks{Australian Centre for Robotic Vision, The Australian National University, Canberra. 
{\tt\small firstname.lastname@anu.edu.au}}
}

\begin{document}

\maketitle
\thispagestyle{empty}
\pagestyle{empty}


\begin{abstract}
	For effective human-robot interaction, it is important that a robotic assistant can forecast the next action a human will consider in a given task. Unfortunately, real-world tasks are often very long, complex, and repetitive; as a result forecasting is not trivial. In this paper, we propose a novel deep recurrent architecture that takes as input features from a two-stream Residual action recognition framework, and learns to estimate the progress of human activities from video sequences -- this surrogate progress estimation task implicitly learns a temporal task grammar with respect to which activities can be localized and forecasted. To learn the task grammar, we propose a stacked LSTM based multi-granularity progress estimation framework that uses a novel cumulative Euclidean loss as objective. To demonstrate the effectiveness of our proposed architecture, we showcase experiments on two challenging robotic assistive tasks, namely (i) assembling an Ikea table from its constituents, and (ii) changing the tires of a car. Our results demonstrate that learning task grammars offers highly discriminative cues improving the forecasting accuracy by more than 9\% over the baseline two-stream forecasting model, while also outperforming other competitive schemes. 
\end{abstract}

\section{Introduction}
Understanding human activities and forecasting the subsequent actions is a fundamental problem in human-robot interaction and co-operation. For example, consider the task of assembling the components of an Ikea table\footnote{We will use the furniture assembly task as a running example to illustrate our scheme, and is also used in our experiments.}. A robot designed to assist in this task must be able to recognize and predict the next human action in the sequential process so that it can hand over a furniture component or a tool. Even for such sequential tasks, real world activities\footnote{Note that we use 'action' to mean an atomic unit and 'activity' to denote a sequence of 'actions'} are often complex and different from each other in very subtle ways (such as~\emph{screwing in} or~\emph{screwing out}), or the same action may have significant variations in their appearances (such as~\emph{assembling on a work bench} or~\emph{assembling on the floor}), may involve hard-to-detect miniature tools (such as~\emph{screws},~\emph{screw drivers},~\emph{hammers}, etc.), may have severe object/body-part occlusions, or may vary in length, complexity, or the speed of the actions. 

A standard way to action forecasting is to model the activity as a Markovian sequence (such as a hidden Markov model) and predict the next action from the previous one~\cite{pentland1999modeling}; however such a scheme loses the global activity context; thereby demanding extra contextual cues such as using object interactions~\cite{kirk2015online}, higher-order MRFs~\cite{chakraborty2014context}, etc. More recently, deep architectures have been suggested for this task such as Ma et al.~\cite{ma2016learning} that proposes to estimate progress of single actions using LSTMs, however it is not clear if the method could be applied to action forecasting or scale to longer videos containing multiple actions.

\begin{figure}[ht]
\includegraphics[width=1\linewidth]{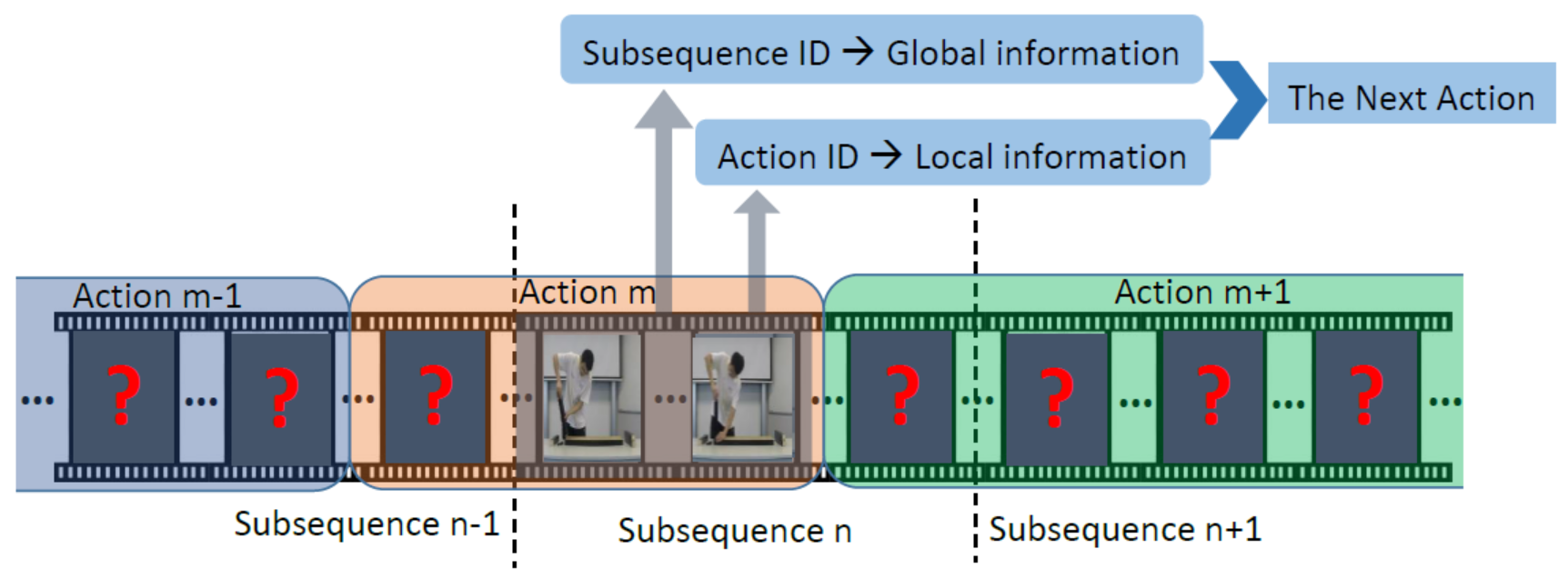}
\caption{An illustration of our proposed action forecasting method. Our scheme takes as input frames from a video subsequence that may contain multiple actions. Our goal is to forecast the next distinct action immediately after the last action in this clip. We propose to use the local information within the given clip (Action ID) and localizing the subsequence progress within a learned task grammar over the full video sequence (via the subsequence ID), to forecast the next action.}
\label{fig:intro}
\end{figure}

In this paper, we propose a novel scheme for action forecasting in sequential tasks by estimating the task progress. Our main intuition is that learning to predict the progress of an activity automatically could learn features that can globally localize the actions within a temporal grammar. Such temporal grammar could then provide a strong global prior over the local Markovian forecasting framework, and could even enable reliable forecasting in the presence of repetitive sub-tasks, or when actions happen out of order. A standard way perhaps to design such a progress estimation framework is to resort to deep recurrent models, such as a long-short term memory modules (LSTM)~\cite{ma2016learning}, or gated recurrent units (GRU)~\cite{Chung14GRU}. Unfortunately, we found that directly using these architectures to regress against continuous scalar activity progress value may lead to sub-optimal results. This is perhaps due to the implausibility of accurately estimating how long a fine-grained action in the task is going to take (e.g., how long a \emph{screwing in} will take), or due to their long range nature (spanning over several minutes). 

In order to mitigate these issues, we propose a novel action forecasting pipeline, which (i) uses the features from a two-stream Residual CNN network for generating local action features, (ii) a stacked LSTM network that takes the features from (i) for estimating the task progress, and (iii) uses the features from (i) and (ii) to forecast the next action. Figure~\ref{fig:intro} shows a schematic illustration of our approach. Our architecture is end-to-end trainable. As noted above, we found it non-trivial to train the LSTM modules for progress estimation by directly regressing to a continuous scalar progress value. To alleviate this issue, we cast the progress estimation problem in a discriminative setup by discretizing it into multiple distinct temporal units, and training the LSTM modules to \emph{classify} a given video sub-sequence of the activity into one of these units.  As using only a single temporal granularity might miss out on short actions, we propose a multi-granularity LSTM framework that is trained to estimate progress at multiple temporal scales. 

A second contribution of our work is the introduction of a novel objective loss for training the LSTM modules for progress estimation. Given that we model the problem in a discriminative setup, it is natural to consider using cross-entropy loss to train such recurrent units. However, we may want to penalize for incorrect progress estimates relative to their proximity to ground truth within the learned temporal grammar. For example, estimation of progress as 1-in-10th should be penalized more than 4-in-10th when the ground truth is 5-in-10th. To incorporate such relative but discrete penalization, we propose to compute the cumulative sum of the progress estimations for sub-sequences, which are then compared to a cumulative sum of the ground truth (which will be a step function) using the Euclidean distance. We train our LSTM modules against the gradients of this loss via back-propagation through time (BPTT).

To demonstrate the effectiveness of our proposed architecture, we present experiments on recorded videos from two tasks, namely (i) the recently introduced \emph{Ikea Furniture Assembly} (IFA) dataset, which consists of video sequences of people assembling pieces of an Ikea Furniture, and (ii) using video sequences depicting \emph{changing car tires} (CCT). For the former, the dataset offers two variants of the task, one on a work bench, where the person is upright in the videos, and the pieces are on a table, while in the second variant, the assembly process happens on  the floor and in which the person could be sitting or standing, thereby introducing occlusions or diversity in the way the pieces are assembled. As for the CCT dataset, the videos are downloaded from the Internet and show significant diversity in the appearances, view points, and order of the actions. Our results (presented in Section~\ref{sec:expts}) demonstrate that our proposed progress estimation pipeline offers superior performance over only using local Markovian architecture, improving the action forecasting performance by nearly 10\%.

%
%
%

\section{Related Works}
\label{sec:work}
In this section, we review the relevant literature associated with problem of human action recognition, forecasting, and progress estimation. 

\subsection{Activity Recognition}
Recognizing human actions from short video clips has been a problem of significant interest in both computer vision and robotics with applications such as robotic learning by human demonstration, human-robot interaction, etc. In the recent years, this problem has advanced from recognizing simple human actions in constrained settings, such as walking, standing, etc.~\cite{schuldt2004recognizing} to highly complex realistic scenarios such as from movies~\cite{laptev2008learning}, TV shows~\cite{patron2012structured}, Internet videos~\cite{lan2012social}, and even real life setups~\cite{soomro2012ucf101}. Such methods could potentially recognize actions in heavy background clutter, occlusions, camera motion, etc. With the recent resurgence of deep learning architectures~\cite{krizhevsky2012imagenet}, the problem of action recognition has seen significant advancements with accuracies plummeting towards near human accuracy. Among the popular models, one model that stands out among others is the two-stream convolutional neural network ~\cite{simonyan2014two,feichtenhofer2016convolutional,feichtenhofer2016spatiotemporal} that pools action predictions from single RGB video frames and short sub-sequences of optical flow respectively. While, we also use a variant of the two-stream model in our architecture, we use it to recognize as well as forecast the subsequent action.

%
\subsection{Action Forecasting}
There have been several recent efforts at forecasting human actions on video sequences over deep models. However, most methods focus on predicting actions from unfinished videos~\cite{li2014prediction,ryoo2011human,cao2013recognize,lan2014hierarchical,kong2016max}. To address this problem, Ryoo et al.~\cite{ryoo2011human} proposes to use integral bag-of-words (IBoW) and dynamic bag-of-words (DBoW), which are two variants of the bag-of-words representations for modeling the temporal evolution of actions. However, the learned model is sensitive to outliers and large variations of appearance in the same class. Using the reconstruction error in a likelihood computation setup to build action models for learning the feature bases is proposed in Cao et al.\cite{cao2013recognize}, a coarse-to-fine hierarchical representation for action prediction is proposed in Lan et al.\cite{lan2014hierarchical}, while the temporal dynamics from observed features are used in Kong et al.~\cite{kong2016max}. Unsurprisingly, CNN based methods have shown state-of-the-art results in several datasets. In Lee et al.~\cite{lee2016human}, spatio-temporal relationships between human objects are captured using a pre-trained CNN and use low level features to represent unfinished human activities. Recurrent deep models have shown promising results in several works. Some notable such works include Ma et al.,  \cite{ma2016learning} in which a ranking loss is applied on the Long Short Term Memory (LSTM) model to retain the monotonicity of the activity progress estimation, while in Becattini et al.\cite{Becattini2017AmIDone}, a fast R-CNN is combined with an LSTM, dubbed ProgressNet, which can not only predict action labels, but can also localize the actions spatio-temporally. In contrast to the above schemes that assume the presence of a single action in the video, our method is significantly different in several ways, namely (i) our scheme assumes long video sequences (5-10 minutes long), (ii) consisting of multiple sequential actions, a setup non-trivial for most modern deep recurrent models. 

\subsection{Activity Progress Estimation}
A third and important component in our architecture is the use of a temporal grammar obtained via learning to estimate activity progress. Modeling activity progress, especially for long and complex videos sequences has been investigated earlier. For example, a ranking loss is applied in Ma et al.~\cite{ma2016learning} to capture the progress of action globally from the the beginning of a video to the current time; this loss is combined with the classification loss to train the networks. Similarly, the progress of activities is predicted using a stacked LSTM in Becattini et al.~\cite{Becattini2017AmIDone}. However, both \cite{ma2016learning} and \cite{Becattini2017AmIDone} assumes a single action, while we envisage learning the progress of an entire task consisting of multiple actions of varying durations.

To the best of our knowledge, it is for the first time that a multi-granularity LSTM network is proposed on two-stream CNN features for action forecasting via learning temporal action grammars for progress estimation. 

\section{Proposed Method}
\label{sec:method}
\begin{figure*}[ht]
	\centering
	\includegraphics[width=16cm,trim=0cm 5cm 6cm 0cm, clip]{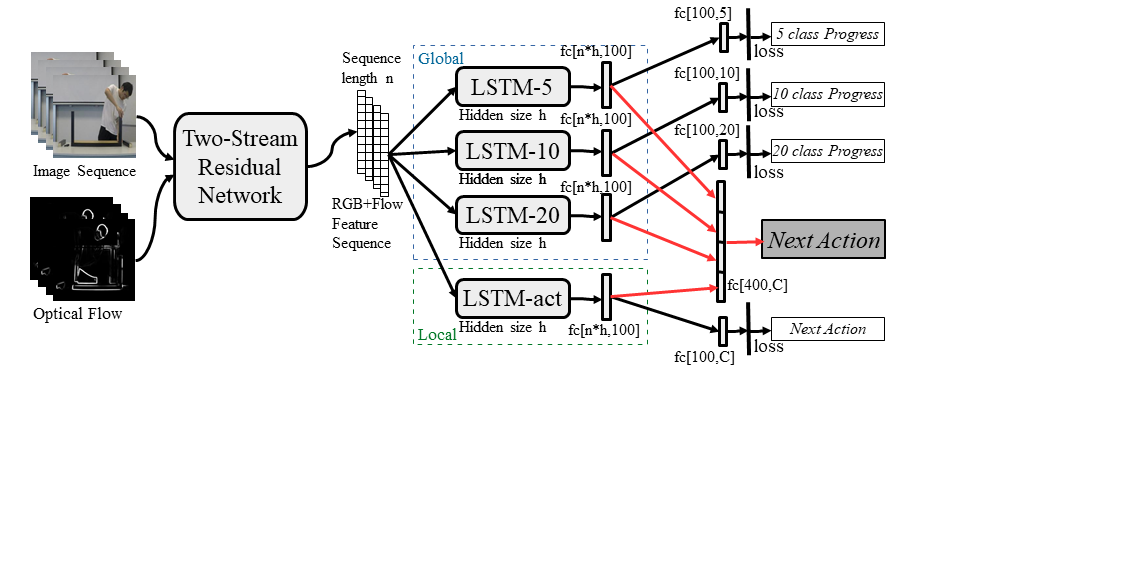}
	\caption{Our full end-to-end learnable action forecasting pipeline. The global progress estimation constituents as well as the local action-to-action forecasting parts are highlighted. LSTM-act denotes our local forecasting module, while LSTM-X denotes learning task grammars by learning progress quantized into X number of distinct units from the start to finish of the task.}
	\label{fig:struct}
\end{figure*}

In this section, we provide details of our proposed action forecasting architecture. First, we introduce our two-stream residual CNN model to forecast actions `locally' i.e., predicting the next action from the current one. Next, we provide details of our stacked LSTM network for learning a task grammar, following which our full progress estimation framework is introduced that predicts progress at multiple temporal scales using a new cumulative progress loss. In Figure~\ref{fig:struct}, we illustrate the complete forecasting pipeline. Before detailing our approach, we introduce our notations and clarify a few realistic assumptions regarding our experimental setup that could help disambiguate our approach described in the sequel. 

\noindent\paragraph{Assumptions}
First, we assume that our inputs are regular RGB video sequences captured by a stationary camera. While, we could use other sensing modalities such as 3D sensors, or stereo cameras, that could potentially improve our accuracy, we decided to use a very general setup of 2D video sequences. This is because, such sequences are available aplenty on Internet websites, and thus can be downloaded easily to train our deep CNN models. For example, the CCT dataset that we use in our experiments is of this nature. We assume the input to our scheme is a sub-sequence of such a long video; the sub-sequence consisting of one or several actions. 

Second, as alluded to above, the actions that we tackle in this paper could be arbitrarily long, involving hard to detect tools or components. Thus, we assume it is implausible to estimate individual action progress. This assumption leads to an ambiguity in our evaluation as it is difficult to say if the forecasted action is a continuation of the last action in the sub-sequence, or not. To resolve this ambiguity, we assume to always forecast an action that is distinct and immediately follows the last one in a given subsequence.

\subsection{Problem Formulation}
\label{sec:problem_formulation}
Let us assume we are given a set $\seqset=\set{S_1, S_2, \cdots, S_n}$ of $n$ video sequences, each $S\in\seqset$ consisting of $K$ actions (including repetitions) from $C$ distinct action classes. We assume that there is an underlying grammar based on which these actions happen, however the grammar is less rigid, allowing flexibility in the occurrences of some of the actions in arbitrary order.\footnote{Concretely, in the furniture assembly task, while the underlying goal is to start from a grammar state where all the furniture pieces are disassembled towards a goal state in which the table is assembled. However, during the assembly process one could first \emph{pick up} a table leg, \emph{attach} it, and then \emph{pick up} another leg, or one could \emph{pick up} two legs, and \emph{attach} them together, but one cannot \emph{attach} legs without first \emph{picking} them up. Having such uncertainties in the order of some elements in the grammar makes dictating one manually difficult, motivating to explore frameworks that could allow learning such task grammars from data.} Further, without loss of generality, let $S = \langle f_1, f_2, \cdots, f_M\rangle$ denote the $M$ ordered frame representations (could be CNN features computed on video frames) in sequence $S$, and let $S'=\langle f_1, f_2, \cdots, f_{M'} \rangle$ be a subsequence of $S$ with $M' < M$ frames. If $y(f_i)$ denotes the action\footnote{This could be a `no-action' background class as well.} associated with frame $f_i$, then going by our second assumption above, our goal is to forecast the action $y(f_{m})$ from $S'$, where for the minimum $m > M'$ such that $y(f_{m})\neq y(f_{M'})$. To avoid the end cases in our evaluation, we assume $y(f_{M'})$ is not the first or the last action in $S$. 

\subsection{Learning Action Features}
The first step in our scheme is to learn useful frame level representations that can model action cues. To extract powerful syntactic and semantic action features, we decided to use the deep features from a CNN. In terms of the architecture of the network, people usually face the trade-off between the depth of the network and the training complexity. Fortunately, \cite{he2016deep} provide enough analysis and solutions for this issue by applying residual mapping instead of unreferenced mapping to address the degradation problem, which keep the depth of the network but with relatively less parameters. Specifically, we apply ResNet-152 model with two-stream fusion, which is mentioned in Section \ref{sec:work}. We use supervised learning strategy on each frame in each sequence in the spatial stream, and 10 stacked optical flow around that frame in the temporal stream.\footnote{Note that we use ResNet-152 as the feature extractor, so we apply supervised learning for action recognition instead of prediction.} At last, we extract the feature from the last fully connected layer of both streams, concatenate them and use as inputs to the subsequent LSTM layers for forecasting.     

\subsection{Local Forecasting Model}
\label{sec:local}
As alluded to above, pooling on tiny action subsequences, while often seen to be useful for action recognition, they may lose the temporal evolution of actions. Such evolutions are important for encoding activity grammars for forecasting (as also demonstrated in our experiments in Table~\ref{table:local+global}). To learn the temporal evolution, we feed the CNN features to a stacked LSTM module, dubbed our `local' forecasting model (see Figure~\ref{fig:struct}). This LSTM sub-network is trained to directly predict the next action from subsequences $S'$ of $S\in\seqset$ (described in Section~\ref{sec:problem_formulation}).

\subsection{Learning Task Grammars}
\label{sec:grammar}
In this section, we detail the core idea of this paper, i.e., learning task grammars using a deep recurrent architecture via estimating the progress of activities in given sub-sequences. Using the notations introduced in Section~\ref{sec:problem_formulation}, a straightforward way for progress estimation is to discretize the length $M$ of sequence $S$ into $N$ distinct intervals, such that for a subsequence $S'$ of length $M'$, the ground truth progress $g_{S'}$ is given by:
\begin{equation}
g_{S'} = \floor{\frac{M'}{M} N},
\label{eq:bin}
\end{equation}
where $\floor{.}$ is the standard \emph{floor} operation. We propose to train a recurrent network against a loss defined on such discretized progress ground truths. In the experiments in the sequel, we explore various possibilities for such recurrent architectures, including LSTMs, RNNs, and GRUs. 

An important ingredient in our scheme is the granularity of the progress interval, i.e., $N$. A bin size of $N=M$, will lead to a continuous progress estimation, which might be ideal. However, such a deep network could be difficult to train as there might not be sufficient motions between frames that they can be mapped to monotonically increasing progress numbers. On the other hand, the other extreme of having $N=1$, is also not ideal for similar reasons. Given that different actions in the task may have different durations to complete, we propose to use a multi-granularity recurrent network that are trained for an $N$ in a set $\mathcal{N}=\set{N_1, N_2, \cdots, N_k}$. For each, $N\in\mathcal{N}$, we model a separate network that forms a parallel progress estimation stream to all other such recurrent networks. Such a multi-granularity LSTM network using $\mathcal{N'} = {5,10,20}$ is shown in Figure~\ref{fig:struct}. 

As alluded to above, we hypothesize that such multi-stream LSTM networks allow learning multiple grammars in a coarse-to-fine setup, at different temporal scales. When trained end-to-end, such multiple networks allow the actions in a given subsequence to be discriminated and localized against each granularity -- the features from each stream when combined provides a global progress of the task even when there are repetitive actions. 

\subsection{Loss Formulations}
\label{sec:loss}
Defining effective loss functions are extremely important when training very deep architectures. As described above, for our task of learning action grammars, one could directly use the standard Euclidean loss on the discretized progress ground truths, i.e., if $\hat{p}_{S'}$ is the predicted progress, then the squared Euclidean loss $\loss_E$ will have the following form:
\begin{equation}
\loss_E(g_{S'}, \hat{g}_{S'}) = \enorm{g_{S'} - \hat{g}_{S'}}^2.
\label{eq:euclid}
\end{equation}

While, this is simple to use, it may not be effective in learning to discriminate various time intervals. This is because, the form of the Euclidean loss as depicted in~\eqref{eq:euclid} penalizes more for intervals towards the end of the sequence in comparison to the beginning. For example, consider the scenario when the ground truth $g_{S'}$ is at 9-in-10ths while the progress estimation $\hat{g}_{S'}$ is at 1-in-10ths. As is clear, this loss will be 7 times more than a loss of $\hat{g}_{S'}$ is 1-in-10ths against $\hat{g}_{S'}$ is 4-in-10ths. Such a bias affects learning the weights in the network leading to sub-optimal network training.

One way to mitigate the above problem is to cast the loss in a discriminative setup by training the LSTMs to predict one of the $N$ discretized classes. This allows including a standard classification loss into the setup via for example a softmax component followed by a cross-entropy loss. However, with such a choice we sacrifice the specifics of the sequential progress task. That is, a cross-entropy loss penalizes all incorrect predictions equally, however, we may better be penalizing the predictions based on some notion of their distance to the ground truth -- as in the case of the Euclidean loss. 

Making both ends meet, we propose a novel cumulative probability loss $\cploss$ by recasting the LSTM progress predictions as discrete probability distributions, whose cumulative distributions are used as the progress representations. Concretely, suppose $v\in\binary{N}$ denotes a one-off vector capturing the progress of a subsequence $S'$ of $S$, whose $p_{S'}$-th dimension (for $p_{S'}$ as defined in~\eqref{eq:bin}) is one. Further, let our LSTM progress estimation module produces as output a vector $\hat{v}\in\reals{N}$, where its $i$-th dimension, $\hat{v}_i$ denotes the classification confidence of the network for the progress to belong to the $i$-th bin. Then, if $\sigma(.)$ denotes element-wise sigmoid, we propose our cumulative probability loss $\cploss$ as:
\begin{equation}
\cploss(v,\hat{v}) =  \enorm{\left(\frac{\sigma\left(\hat{v}\right)}{\sigma\left(\hat{v}\right)^T\one}-v\right)M}^2,
\label{eq:cploss}
\end{equation} 
where the $\sigma$ maps each dimension of $\hat{v}$ to the range $(0,1)$, and $M$ is an upper-triangular matrix with all ones over the main diagonal. The first term in~\eqref{eq:cploss} converts the vector $\hat{v}$ to a discrete probability vector, while using $M$ produces a cumulative distribution function. Note that such a loss easily avoids the problems with the scalar Euclidean and cross-entropy losses by balancing the weights associated with each progress bin using the cumulative distribution matrix $M$.  The $k$-th dimension of the gradient of this loss is given by:
\begin{equation}
\nabla_{\hat{v}_k}\cploss=2 h(\hat{v}_k) \frac{\sum_{i=k+1}^N\sigma(\hat{v}_i)}{\left(\sigma(\hat{v})^T\one\right)^2}\nabla_{\hat{v}_k}\sigma(\hat{v}_k),
\end{equation}
where $h(\hat{v}_k)$ is the $k$-th dimension of the component inside the squared-Euclidean term in~\eqref{eq:cploss}.

\subsection{Local + Global End-to-End Model} 
To use the complimentary nature of the `local' action forecasting model with the `global' task grammars, we propose to combine the `local' and `global' model to build a joint action forecasting network. Our combined model (as depicted in Figure~\ref{fig:struct}) uses two sets of losses on the LSTM progress estimation modules, namely (i) we optimize for a task progress loss as dictated by the $\cploss$ and (ii) an action forecasting loss captured by cross-entropy over a softmax-ed prediction score over the next action. We combine the intermediate features from these modules along with the features from the `local' forecasting module to generate a combined action feature that is trained for predicting the next action.

\section{Experiments}
\label{sec:expts}
\begin{figure}[htbp]
\centering
\includegraphics[width=0.8\linewidth]{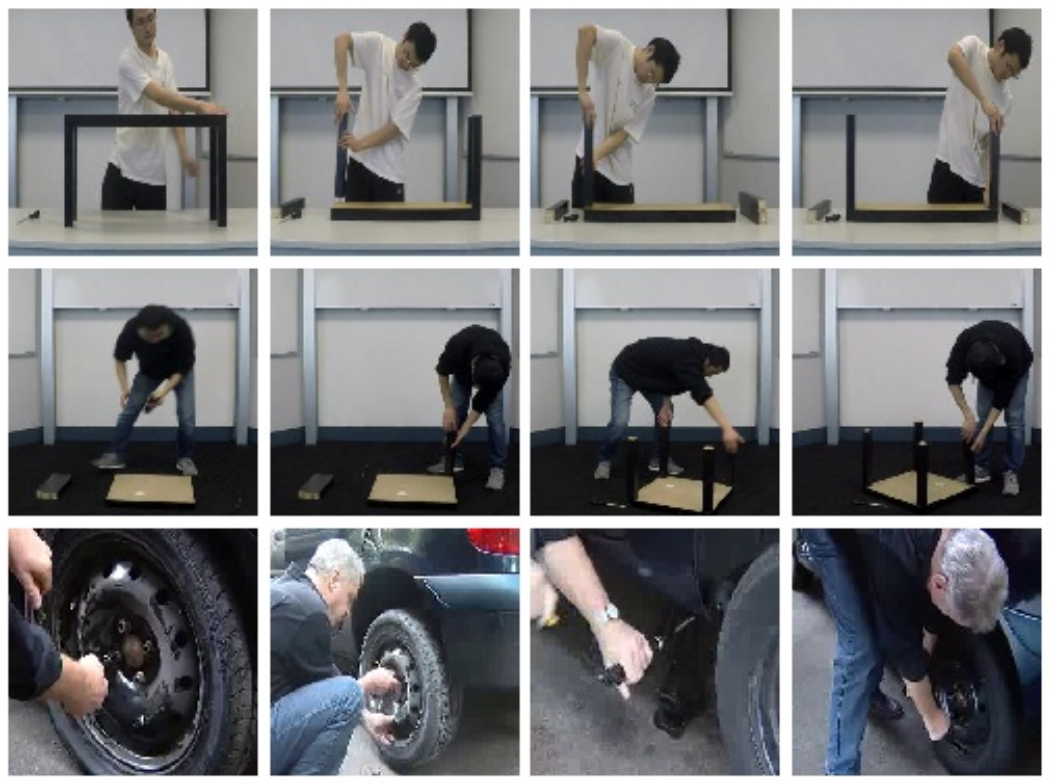}
\caption{Overview of dataset: first row is IFA-Bench, second row is IFA-Ground, and third row is Changing Tire Dataset.}
\label{fig:dataset}
\end{figure}

In this section, we provide experiments validating the performance of our proposed schemes. As the primary goal of this paper is to use action forecasting for human-robot interaction, we use datasets that are aligned towards such a scenario. To this end, we use the recently introduced Ikea Furniture Assembly (IFA) dataset~\cite{Toyer2017DMM} that consists of videos sequences of multiple people assembling small pieces of an Ikea table. In order to further validate the applicability of our method on more general scenarios, we also report experiments on videos from the~\emph{Instruction Videos} dataset~\cite{Alayrac16unsupervised}. Specifically, we use the subset of videos associated with \emph{changing the tires of a car}, which we call the \emph{Changing Car Tires} (CCT) subset. Sample frames from these datasets are provided in Figure~\ref{fig:dataset}.

\subsection{Datasets}
\noindent\paragraph{Ikea Furniture Assembly Dataset~\cite{Toyer2017DMM}} consists of 101 videos in total, each video about 5-7 minutes long containing a single person assembling an Ikea furniture, captured by a stationary GoPro camera in HD quality. In each video, the person assembles and then disassembles the table following some fixed, but flexible  procedures. The table is assembled by spinning each leg into the corners of the table-top. The legs and table-top are connected by a double-head screw. Each participant starts with a table-top and four legs on the side, then assembling the table, and then disassembling the same table to reach the initial state. Usually, people assemble and disassemble legs one by one. Temporal action annotation is available for this dataset. There are totally 13 classes of actions in this task, namely \emph{pick leg}, \emph{attach leg X}, \emph{detach leg X}, \emph{flip table}, \emph{spin in}, \emph{spin out} and a \emph{null action}, where $X$ is a number in 1--4. The \emph{Null action} indicates that the current time interval does not belong to any other action class; such \emph{null actions} appear between useful actions. In our experiments, we remove \emph{null action} as it does not correspond to any meaningful progress in the task grammar. The frame index of each of frame is also provided with the dataset, which we use to estimate or generate the ground truth. 

Half of the videos in this dataset have the furniture assembly task on a work bench, while the other half is on the floor. The former category consists of upright assemblers and the furniture pieces and the tools are on the work bench, while for the latter, the assembler has to bend or sit on the floor for the task, making the action recognition much more challenging due to  irregular human poses and body-part occlusions. In our experiments, we separate the two varieties as separate sub-datasets, dubbed Ikea-Bench and Ikea-Ground. 


\noindent\paragraph{Changing Car Tires~\cite{Alayrac16unsupervised} Dataset} consists of 30 videos with 11 distinct actions. Each sequence shows a process of how people change the tire of a car. Differently from IFA, in this dataset, each action only appears once in each video and follows a fixed order. Further, not all the actions may happen in every video, making it quite challenging. Sometimes the actor skips some actions for changing tires. There also appears several null/noisy background actions between the tire-changing process, such as chatting or walking around. All null actions are removed in our experiments.

\subsection{Deep Models}
All four LSTM blocks described in~\ref{sec:grammar} are stacks of two LSTMs layers with 32 hidden units in each layer. A dropout layer (with ratio 0.2) is added on the output of the first LSTM layer to mitigate potential over-fitting. A fully-connected (fc) layer follows each LSTM block, taking all the hidden states as input and generating feature vectors with length 100. For the global progress pipelines (LSTM-5,10,20), another fc layer follows the previous fc layer, and generates progress estimation in different granularity (see Figure~\ref{fig:struct}). For the local action prediction pipeline (LSTM-act), another fc layer is attached to generate the classification result for the forecasted action. In the local+global pipeline, which is denoted by red arrows in Figure~\ref{fig:struct}, the outputs of the four fc layers are concatenated into a long vector with length 400, and is fed into a larger fc layer to train for the action forecasting. Adam optimizer~\cite{DBLP:journals/corr/KingmaB14} with initial learning rate of $1e^{-4}$, (0.9,0.999) $\beta$s and $1e^{-8}$ epsilon are used to train all local, global, and combined models. 

\subsection{Data Splits}
There are 14 actors in IFA dataset. For both Ikea-Bench and Ikea-Ground, we use videos from 11 actors for training and validation, and the rest for testing. Specifically, we use all videos with assembler ids 9 and 11 for testing Ikea-Bench, and ids 9 and 13 for testing Ikea-Ground. There are a total of 8 test sequences, 5 on the work bench and 3 on the floor. The CCT dataset contains 30 labeled Youtube videos, we select label 2,6,9,13,17,25 as the testing set, and we train on the rest.

\subsection{Evaluation Protocols}
For both the datasets, we report average forecasting accuracy, and the respective mean precision, mean recall, and the confusion matrices against the ground truth forecast.Mean precision and mean recall are obtained by averaging precision and recall for each class.

\subsection{Experimental Setup}
\subsubsection{Task Progress Estimation}
An important choice to be made in our evaluation is the type of recurrent architecture to be used for activity progress estimation. In this section, we empirically evaluate potential models such as LSTMs, RNNs and GRUs. To show the performance on our proposed cumulative probability loss (CPloss), we compare three loss functions for the progress estimation task, namely (i) the L2 loss for regression, (ii) Cross-entropy loss for classification and (iii) the proposed cumulated probability loss (CPLoss). We run regression for progress estimation task as a base line model. We run the experiments to investigate the accuracy on different time interval lengths (5,10,20), sub-sequence lengths (5, 10, 20 frames) and different loss functions. 

\begin{figure}[htbp]
\centering
\includegraphics[width=1.0\linewidth]{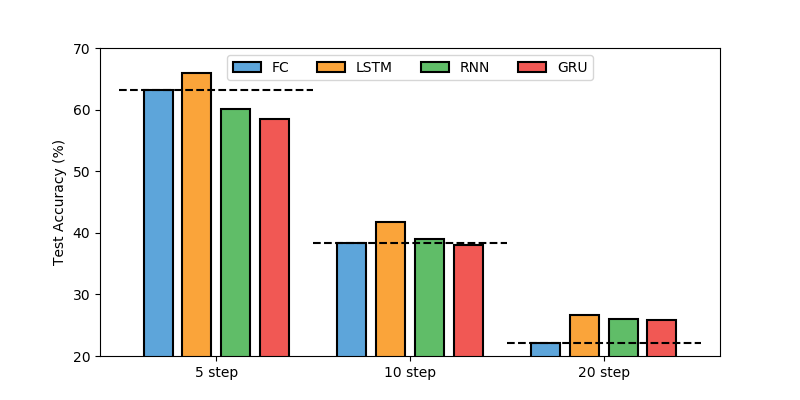}
\caption{Models vs progress steps on IFA-Bench with sequence length = 10 frames}
\label{fig:model_compare}
\end{figure}




\begin{table}
  \centering
  \begin{tabular}{|c|c c c c|}
    \hline
    Length of sequence & 5 & 10 & 20 & 50 \\
    \hline
    5 steps & 61.5\% & 61.8\% & \textbf{63.4\%} & 58.2\%  \\
    10 steps & 45.1\% & \textbf{48.2\%} & 46.1\% & 38.9\% \\
    20 steps & 27.8\% & 28.8\% & 25.8\% & \textbf{29.1\%} \\
    \hline
  \end{tabular}
  \caption{Progress step vs length of sequence for global progress estimation for LSTM model on IFA-Bench dataset}
  \label{tab:1}
\end{table}

\begin{table}
  \centering
  \begin{tabular}{|c|c|c c c|}
    \hline
     & Progress steps & 5 steps & 10 steps & 20 steps \\
    \hline 
    \multirow{3}{*}{Ikea-Bench} & L2 Loss & 48.7\% & 33.1\% & 17.6\% \\
                                  & Cross-Entropy Loss & 61.8\% & 48.2\% & \textbf{28.8\%} \\
                                  & \textbf{CPLoss (ours)} & \textbf{64.3\%} & \textbf{50.0\%} & 25.6\% \\
    \hline 
    \multirow{3}{*}{Ikea-Ground} & L2 Loss & 34.2\% & 20.2\% & 10.5\% \\
                                   & Cross-Entropy Loss & \textbf{47.8\%} & \textbf{29.9\%} & 15.9\% \\
                                   & \textbf{CPLoss (ours)} & 47.0\% & 28.3\% & \textbf{18.1\%} \\
    \hline
  \end{tabular}
  \caption{Progress step vs Loss functions for LSTM model on IkeaFA with sequence length 10}
  \label{tab:loss}
\end{table}

\subsubsection{Local activity prediction}
This section explains the experimental setup on local activity prediction, which is the pipeline containing LSTM-act block in~\ref{sec:local}. In a specific human activity, it is natural that some actions have shorter duration, and the other span for longer time. For example, in our IFA dataset some actions like \emph{picking table legs} are really short, and actions like \emph{spinning legs} may occupy most of the video duration. In this case, if we uniformly or randomly select training sequences, those rare actions have very limited probability to be chosen and be trained, which influence the performance of our model. To avoid such cases, we adjust the probability of sampling sequences in our training, and ensure that every action label has similar frequency to be trained. For testing, we uniformly select all action sub-sequences for every video in test set.\\

\begin{table}
\centering
\begin{tabular}{|c|c c c c|}
\hline
length of sequence & 5 & 10 & 20 & 50 \\
\hline
IkeaFA-Bench & 46.4\% & 44.7\% & 48.4\% & \textbf{51.7}\% \\
IkeaFA-Ground & 46.9\% & 37.4\% & \textbf{47.7}\% & 43.2\% \\
\hline
\end{tabular}
\caption{Local LSTM model: IkeaFA dataset vs length of sequence for LSTM}
\label{tab:local}
\end{table}


\begin{table*}[ht]
\centering
\begin{tabular}{|c|c|c c c|c c c|}
\hline 
& & \multicolumn{3}{c|}{Cross-Entropy Loss} & \multicolumn{3}{c|}{CPLoss} \\
& & accuracy & mean precision & mean recall & accuracy & mean precision & mean recall \\
\hline
\multirow{4}{*}{IFA-Bench}
&local    & 47.1\% & 47.0\% & 52.4\% & 47.1\% & \textbf{47.0}\% & 52.4\% \\
&      +5 & 49.9\% & \textbf{48.6}\% & 57.3\% & 47.4\% & 45.3\% & 55.0\% \\
&   +5+10 & 49.6\% & 45.5\% & 57.2\% & \textbf{53.0}\% & 46.7\% & \textbf{59.5}\% \\
&+5+10+20 & \textbf{56.7}\% & 46.5\% & \textbf{60.2}\% & 50.2\% & 45.4\% & 58.1\% \\
\hline
\multirow{4}{*}{IFA-Ground}
&local    & 40.8\% & 28.6\% & \textbf{34.5}\% & 40.8\% & 28.6\% & 34.5\% \\
&      +5 & 44.2\% & 28.6\% & 33.8\% & 44.2\% & 29.4\% & 33.3\% \\
&   +5+10 & \textbf{46.3}\% & \textbf{28.9}\% & 34.2\% & 44.3\% & \textbf{29.7}\% & 33.7\% \\
&+5+10+20 & 45.6\% & 28.2\% & 33.9\% & \textbf{50.2}\% & 29.5\% & \textbf{35.7}\% \\
\hline
\multirow{4}{*}{Change Tire}
&local    & 32.3\% & 26.5\% & \textbf{29.3}\% & 32.3\% & 26.5\% & \textbf{29.3}\% \\
&      +5 & 33.1\% & 24.9\% & 28.4\% & 32.7\% & 25.4\% & 29.2\% \\
&   +5+10 & \textbf{33.5}\% & 26.3\% & 28.0\% & 32.6\% & 25.3\% & 28.2\% \\
&+5+10+20 & 33.2\% & \textbf{27.4}\% & 28.2\% & \textbf{34.7}\% & \textbf{27.2}\% & 29.2\% \\
\hline
\end{tabular}
\caption{Evaluation of Local + Global action prediction model. We compare different model combinations.}
\label{table:local+global}
\end{table*}


\subsection{Combine local with global}
Considering the performance of both global progress estimation model and local action prediction model and time efficiency, we select length of sequence to be 10 in our final experiments. In order to evaluate the proposed CPLoss with traditional cross-entropy loss, we run separate experiments on each loss function. In~\ref{sec:loss}, the column `Cross-Entropy Loss' and `CPLoss' indicate the loss function for global progress estimation model, while the loss for training local action estimation model is always cross-entropy loss. In order to show the effect of combined granularity, we evaluate four models in this experiments: local action prediction model as the baseline, local model with 5-class progress model (denoted as +5), local model with 5-class and 10-class progress model (denoted as +5+10), and local model with all three progress model (denoted as +5+10+20). The last configuration is the exact model shown in Figure~\ref{fig:struct}. 


\subsection{Experimental Results}
Figure~\ref{fig:model_compare} shows the comparison of LSTM, RNN, and GRU on the progress estimation task with sequence length as 10 frames. Notably, when training, we use subsequences of length $\ell$, where the beginning of the sequences need not be aligned with the start of the sequences. Our goal is to use such random start subsequences as augmented data for effective training. That is, starting from frame, say $i$ in the progress window $j$ consisting of $t$ frames, we select $\ell$ frames all within this $t$ subset, but in order and equi-spaced. Then, this subsequence is used to train the progress network for forecasting. In Figure~\ref{fig:model_compare}, we also report results when using average pooling (FC) instead of the recurrent models. Overall, LSTM  outperforms RNN, GRU, and FC on all 5, 10, and 20 interval progress estimations. On different sequence lengths (such as 20 and 50 frames), LSTM also appears to be the proper choice of recurrent architecture in our experiments. In Table~\ref{tab:1}, we report results on the Ikea-Bench dataset on the progress estimation subtask for increasing length of training subsequences using LSTM models for the deep recurrent architecture. Table~\ref{tab:1} shows that increasing the length $\ell$ improves the forecasting accuracy, and so does decreasing the size of the progress window size (to 20 windows from 10). From the table, it is clear that decreasing the step sizes decreases the accuracy, as recognizing actions from finely granular clips is challenging. Increasing the sequence lengths improves accuracy, which is expected.

In Table~\ref{tab:loss}, we report the performance against several loss functions for different sequence segment sizes (steps). While, on Ikea-Bench, the CP loss performs the best, we find that on IFA-Ground, it performs similar to Cross-entropy, perhaps because of the  features may not characterize the actions effectively due to the large variations in the appearances. In Table~\ref{tab:local}, we analyze the performance of the local LSTM network for increasing subsequence lengths. As is clear, the trend shows that performance increases when the sequence length increases.

Finally, in Table~\ref{table:local+global}, we compare all the different setups against each other on all the three datasets. As is clear, the proposed local + global model obtains superior forecasting accuracy, and shows about 9.6\% improvement on IFA-Bench dataset with cross-entropy loss. For IFA-Bench, we obtained 9.6\% improvement with cross-entropy loss and a 5.9\% improvement with CPLoss. For IFA-Ground, we obtained 4.8\% improvement by cross-entropy loss and 9.4\% improvement by CPLoss. On Changing Tire dataset, the improvement is not that high, but we still obtained 1.2\% on cross-entropy loss and 2.4\% on CPLoss. Overall, it appears that our CP loss consistently shows superior performance against the baselines when combining the local and global models together, however the trend between cross-entropy and the CPloss varies depending on the complexities of the actions -- that is, if the action sequences are less challenging, CPloss performs better, while in the other case, using the cross-entropy classification turns suitable. 


\section{Conclusion}
In this paper, we introduced a novel approach to human action forecasting in sequential tasks by learning temporal task grammars. Specifically, we used the features from a standard two-stream residual network for action forecasting to train a multi-granularity LSTM network for action progress estimation, which implicitly learns the task grammars. This grammar is then used to localize actions, and subsequently used to forecast the next action to a given subsequence. We proposed a multi-stream LSTM network with several different potential loss functions that capture the activity progress. Our experiments on two datasets, closely related to human-robot interaction scenarios, demonstrate the effectiveness of our proposed method, showcasing superior accuracy of our approach when combining the local action forecasting with a global learned temporal grammar.



\addtolength{\textheight}{-12cm}   

{\small
\bibliographystyle{IEEEtran}
\bibliography{bib}}

\end{document}